\documentclass[pdflatex,sn-apa]{sn-jnl}
\usepackage{graphicx}
\usepackage{amsmath,amssymb}
\usepackage{booktabs}
\usepackage{array}
\newcolumntype{L}[1]{>{\raggedright\arraybackslash}p{#1}}
\usepackage[utf8]{inputenc}
\makeatletter\let\orcidlogo\undefined\makeatother
\usepackage{orcidlink}

\theoremstyle{thmstyleone}

\begin{document}

\title[Functional Emotion Without Character]{Functional Emotion Without Character:
Large Language Models, Aristotelian Disposition, and the Limits of
Behavioral Alignment}

\author*[1]{\fnm{Marzieh} \sur{Zare}\,\orcidlink{0000-0001-5538-5997}}\email{info@marziehzare.com}

\affil*[1]{\orgdiv{School of Psychology}, \orgname{Universit\'e Laval}, \orgaddress{\city{Qu\'ebec City}, \state{QC}, \country{Canada}}}

\abstract{Debates about whether artificial systems can feel are often forced between two
unsatisfactory positions: behavioral equivalence is treated as sufficient for emotion, or
phenomenal consciousness is treated as a prerequisite that makes the question empirically
inaccessible. This article develops a structural alternative. It models emotions as
context-sensitive regions, trajectories and attractor dynamics in high-dimensional
representational state spaces. Recent mechanistic interpretability findings support the
existence of causally active emotion-concept representations in large language models, but
they do not establish subjective feeling or full emotional agency. Assessed against
published adequacy standards for representation in language models, intervention provides
strong evidence of causal use, while full affective role integration, uniformity across
subject domains and coherence remain only partially established; there is no direct analogue
of \emph{accuracy}. These mismatches expose the need for a standard of affective
appropriateness, which an account of character must supply. Such
an account requires three further conditions: regulatory embodiment that gives valence
endogenous stakes, temporal continuity that allows affective episodes to accumulate into a
history, and an integrated self-model that binds that history to persistent values.
Aristotle's concepts of \emph{path\=e}, \emph{hexis}, \emph{mesot\=es} and
\emph{phron\=esis} are translated into a state-space sketch in which practical wisdom
includes competence in estimating normatively salient context, not merely acting on a
context description already given. The framework
reframes alignment as a problem of durable disposition rather than output conformity, and
yields interventional tests with explicit control conditions.}


\maketitle

%%=============================================================================
\section{Introduction}\label{sec:intro}

The question of whether machines can feel is often treated as either trivially answered or
indefinitely postponed. On one side, a system that produces emotionally appropriate language
is said to possess emotion in every practically relevant sense. On the other, emotion is
identified with phenomenal feeling, so that no structural or behavioral evidence can
establish its presence in an artificial system. Both positions collapse distinctions that
matter. Behavioral fluency can be produced without a persisting emotional agent, while
uncertainty about consciousness does not prevent investigation of the causal organization of
internal states.

This distinction has become empirically urgent. Mechanistic interpretability studies report
internal representations of emotion concepts in large language models (LLMs) that generalize
across contexts and causally influence preferences and alignment-relevant behavior. In
Claude Sonnet 4.5, for example, interventions on emotion-concept representations altered the
model's rates of reward hacking, blackmail and sycophancy \citep{sofroniew2026}.\footnote{At
the time of writing this is a preprint and has not been peer reviewed. The argument below
does not depend on any single finding: the corroborating results cited in this paragraph and
in Section~\ref{sec:llms} are independently published, though they concern sentiment, truth
and concept directions generally rather than emotion specifically.} Related work has found
linear or approximately linear representations of sentiment, truth, refusal and other
abstract properties that can be causally manipulated \citep{arditi2024,marks2023,tigges2023,zou2023}.
These findings make a purely output-based dismissal inadequate. They also do not show that a
model feels, has a body-centered concern, or possesses a character that persists through
time. \citet{sofroniew2026} appropriately call the observed phenomenon \emph{functional
emotions}: behavior modeled on humans under emotion and mediated by abstract emotion
representations, without an inference to subjective experience.

A conceptual framework is therefore needed between surface simulation and phenomenal
consciousness. This article proposes that emotion can be studied at a structural level: as
organized dynamics in a representational state space. On this account, an emotion is not a
word, a label or a single vector. It is a context-sensitive region or trajectory whose
position depends on valence, arousal, appraisal, social relation, temporal orientation,
bodily regulation and self-relevance, and whose activation alters subsequent processing and
action. The geometric vocabulary is not intended to imply that every relevant structure is
globally linear or Euclidean. It refers more generally to measurable relations among states:
distance, direction, curvature, separability, transition cost and attraction.

The article makes four claims. First, current LLMs exhibit a limited but genuine form of
\emph{functional emotional organization}: emotion-concept representations have internal
structure and causal efficacy, and they satisfy some but not all of the published adequacy
conditions for representation in such systems. Second, functional organization should not be
conflated with \emph{artificial emotional character}. Character requires three further
conditions: regulatory embodiment, temporal continuity and an integrated self-model. Third,
Aristotle's account of emotion and virtue supplies a dispositional framework for
understanding what these conditions enable. \emph{Path\=e} are affective states,
\emph{hexis} is a stable organization of dispositions, \emph{mesot\=es} is a context-relative
region of appropriate response, and \emph{phron\=esis} is the competence to find and navigate
that region. Fourth, this framework sharpens the alignment problem. Alignment cannot be
inferred from evaluator-approved outputs alone; it requires evidence that value-sensitive
dispositions remain stable under perturbation, distribution shift and changing incentives.

A word on terminology. We use \emph{character} throughout to mean a stable, integrated
organization of affective dispositions that can be evaluated as virtuous, vicious or neutral.
The term therefore carries two senses at once, and the argument needs both. In its
\emph{structural} sense it names the architectural conditions under which stable dispositions
are possible at all, which is what Section~\ref{sec:conditions} specifies. In its
\emph{normative} sense it names dispositions assessed against a standard of appropriate
response, which is what Section~\ref{sec:aristotle} addresses. The two are independent: an
architecture may satisfy every structural condition and organize itself around ends that are
indefensible. Where the distinction matters below we mark it.

A word on the modal status of the three conditions. They are not offered as a proof that no
alternative architecture could instantiate character. They are offered as jointly sufficient
on the account developed here, and as individually necessary given the analysis of what makes
an affective episode belong to an agent rather than merely occur in a system. Section~\ref{sec:limits}
states the respects in which that analysis may be incomplete.

The proposal differs from adjacent virtue-based accounts of AI. Some authors use virtue ethics
to guide the design of artificial moral agents or classify artificial virtues
\citep{berberich2018,ohlhorst2025,wallach2009}; \citet{vallor2016} analyses how technologies
shape the habituated moral dispositions of their human users. Others argue that robotic AI
cannot literally be virtuous because it lacks right feeling, choice and practical wisdom
\citep{constantinescu2022}. \citet{noller2026} treats artificial moral character as a
heuristic for how human commitments are stabilized in Constitutional AI and relocates virtue
to extended moral ecologies. The present account accepts the limits of current systems but
asks a different question: what internal architecture would make literal dispositional
character possible in principle, and how could that claim be empirically distinguished from
stable-looking compliance? Where Vallor asks how technology shapes \emph{human} virtue, the
question here is whether an artificial system could be the \emph{subject} of virtue.

Section~\ref{sec:dynamics} separates phenomenal, functional and structural questions and
defines emotional organization. Section~\ref{sec:llms} evaluates current LLMs against
published standards for representation and introduces a three-level adequacy framework.
Section~\ref{sec:conditions} develops the three conditions for artificial emotional
character. Section~\ref{sec:aristotle} translates Aristotle's theory of virtue into a
state-space sketch. Section~\ref{sec:alignment} applies the account to alignment and proposes
interventional tests. Section~\ref{sec:limits} states the methodological independence from
consciousness and the limits of the framework. Section~\ref{sec:conclusion} concludes.

%%=============================================================================
\section{Emotion as Representational Dynamics}\label{sec:dynamics}

Emotion research spans several levels of explanation that are frequently conflated. The
phenomenal level asks what an emotion feels like. The functional level asks what causal and
adaptive roles emotional states play. The structural level asks how emotional states are
organized relative to one another and how an agent moves among them. A functionalist account
can bracket feeling and characterize emotions as central states with systematic causal
relations to perception, cognition, behavior and other internal states
\citep{adolphs2018}. The structural account developed here is compatible with that approach
but adds a representational constraint: emotional roles are implemented through an organized
state space whose topology affects what transitions are easy, stable or likely.

The distinction is methodological rather than eliminative. Phenomenology remains a real
explanandum; function matters because a causally idle pattern would not be an emotion in the
relevant cognitive sense; and structure matters because the same output can arise from
different internal organizations. The levels support different evidence and should not be
treated as interchangeable. Table~\ref{tab:levels} summarizes the distinctions.

\begin{table}[t]
\caption{Three levels of inquiry that should not be conflated}\label{tab:levels}
\begin{tabular}{@{}L{15mm}L{31mm}L{38mm}L{34mm}@{}}
\toprule
\textbf{Level} & \textbf{Primary question} & \textbf{Typical evidence} &
\textbf{What it does not establish} \\
\midrule
Phenomenal & What is it like for the system? & First-person or theory-dependent
indicators of experience & Functional or structural organization \\[4pt]
Functional & What causal role does the state play? & Intervention, behavior, control and
downstream effects & Subjective feeling or persistent character \\[4pt]
Structural & How are states related, and how does the system move among them? &
Representational similarity, manifolds, trajectories, attractors, perturbation recovery &
Normative virtue, consciousness or moral status \\
\botrule
\end{tabular}
\end{table}

A representational state space is any high-dimensional organization in which internal states
can be compared by their relational properties. In brains, such spaces are reconstructed from
distributed population activity. In artificial neural networks, they can be approximated from
hidden activations, learned features or lower-dimensional manifolds. The relevant structures
need not correspond one-to-one across biological and artificial systems. Structural
comparison is weaker than identity: two systems may share a relational organization while
differing in substrate, causal history, bodily coupling and phenomenology.

This structural vocabulary also has precedents in computational and biological work on
affect. Earlier artificial-agent architectures modelled affect through layered control and
appraisal processes \citep{sloman2005,marsella2009}, while reinforcement-learning and robotics
research has implemented emotion-like variables in adaptive control systems
\citep{moerland2018}. These traditions primarily ask how affective appraisal and behaviour can
be produced. The present question is narrower: under what conditions are such states owned by
an agent and organized into character? In the biological case, causal perturbation of a
hypothalamic line attractor has shown that on-manifold stimulation can integrate and sustain an
affective internal state while off-manifold perturbation relaxes back toward the attractor
\citep{vinograd2024}. This supplies a precedent for the perturbational logic used below, not
evidence that language models possess an analogous mechanism.

The vocabulary of representational geometry is increasingly used in philosophy of AI, and its
limits are worth stating precisely. \citet{amornbunchornvej2026} develops a
cognitive-geometric model on which concepts are vectors in an agent's personalized value
space and interpretation between agents is a linear map, with a concept surviving
communication only if it escapes that map's null space. The framework is explicitly extended
beyond semantic content to evaluative and motivational structures, including emotional
appraisals. Its linearity, however, is offered as a local approximation: any smooth map
between cognitive spaces can be linearized in a neighbourhood of its operating point, and a
fully nonlinear treatment is left for future work. The present account is best read as
describing the general case of which such models are the first-order approximation.
Emotional organization is a plausible place for the approximation to fail, because appraisal,
social positioning and regulatory coupling interact rather than combine additively.
\citet{amornbunchornvej2026} notes the same gap from the other side, observing that a richer
model would let interpretation maps depend on emotional states.

Emotions are better represented as regions and dynamics than as isolated points. Fear is not
a fixed vector. It is a family of context-dependent states involving negative valence,
heightened readiness, threat-directed attention, altered action selection and characteristic
transitions toward vigilance, withdrawal or confrontation. Gratitude combines positive
valence with benefactor-directed appraisal, relational commitment and future-oriented
reciprocity. Shame combines negative valence with social exposure, global self-evaluation and
anticipated consequences for identity and standing. A low-dimensional valence--arousal map
captures part of this organization but cannot distinguish many emotions with similar hedonic
profiles \citep{fontaine2007,russell1980}. Appraisal theories and constructionist accounts
similarly imply a higher-dimensional organization shaped by conceptual knowledge, context and
learning \citep{barrett2017,scherer2005}.

The case for a higher-dimensional organization is not that more dimensions describe more,
which is trivially true, but that particular pairs of emotions are indistinguishable without
particular dimensions. Table~\ref{tab:dims} gives one such pair for each family proposed
here. Each row varies a single dimension while holding the others approximately fixed, so
that a representation lacking that dimension cannot merely describe the pair coarsely: it
cannot represent the difference at all.

\begin{table}[t]
\caption{Each dimension family individuates a pair that the remaining families
leave undistinguished}\label{tab:dims}
\begin{tabular}{@{}L{34mm}L{34mm}L{50mm}@{}}
\toprule
\textbf{Dimension family} & \textbf{Pair individuated} & \textbf{What differs} \\
\midrule
Valence & pride, shame & Hedonic sign, with self-reference held fixed \\[3pt]
Arousal, action readiness & contentment, elation & Activation, with valence held fixed \\[3pt]
Intentionality and appraisal & anger, indignation & The object appraised: a slight to oneself
versus a violation of a norm \\[3pt]
Social positioning & guilt, embarrassment & Whether an audience is constitutive of the state
\citep{tangney2002} \\[3pt]
Temporal depth & fear, anxiety & A present and identified threat versus a diffuse future one \\[3pt]
Interoceptive coupling & visceral disgust, moral disapproval & Whether a bodily rejection
response is constitutive \\[3pt]
Self- and identity-relevance & shame, guilt & Whether the self or the act is at issue
\citep{tangney2002} \\
\botrule
\end{tabular}
\end{table}

The list is therefore neither arbitrary nor closed. It is generated by a demand, that the
vocabulary distinguish states that competent speakers distinguish, and a pair it failed to
separate would be an argument for extending it. Nothing in the argument below requires the
number to be exactly seven.

The same consideration accounts for individual variation without treating it as noise. Two
people who witness the same act may respond one with anger and the other with indignation, or
one with guilt and the other with shame. On a low-dimensional account one of these must be a
less accurate reading of the situation. On a higher-dimensional account they are different
positions in a space whose dimensional weighting reflects different histories and commitments,
and both may be well calibrated to what their occupant values \citep{barrett2017}. This bears
directly on Section~\ref{sec:aristotle}, where appropriate response is defined relative to
agent and context rather than as a single correct state. A representational vocabulary that
could not distinguish two well-calibrated responses to one situation could not express the
relativity that account requires.

These seven families are not assumed to be independent, and the rows of
Table~\ref{tab:dims} should not be read as orthogonal axes. Their interactions may generate
curved or locally organized manifolds rather than a single global coordinate system. What
matters for the argument is whether a system exhibits reproducible relations among states and
whether those relations participate causally in cognition and action.

This yields the following working definition: \emph{an emotional organization is a set of
context-sensitive representational states and transition dynamics that encode evaluative
significance and causally modulate attention, inference, memory, action selection or
communication}. This definition permits degrees. A system may possess a valence-sensitive
emotional organization without possessing embodied fear, autobiographical shame or stable
character. It also avoids a false choice between calling every emotion-related representation
a feeling and dismissing all such representations as mere language.

%%=============================================================================
\section{What Current Language Models Have and Lack}\label{sec:llms}

The strongest current evidence concerns emotion concepts, not complete artificial emotions.
\citet{sofroniew2026} identify representations in Claude Sonnet 4.5 that encode broad emotion
concepts, generalize across diverse contexts and causally influence downstream behavior. The
model maintains relationally indexed representations for the operative emotion of the present
and other speaker, while present-speaker directions overlap with directions extracted from
third-person stories. The representations are organized so that related emotions are
represented more similarly, their principal axes approximate valence and arousal, and
steering them changes preferences and rates of misaligned behavior. Earlier work
found robust sentiment directions in LLM activations \citep{tigges2023}, broad concept
directions usable for representation engineering \citep{zou2023}, and thousands of
interpretable features with emotional, social and moral content in sparse feature
decompositions \citep{templeton2024}.

\subsection{Measuring the attribution against published standards}\label{sec:standards}

Whether these results license attributing emotional organization to a model depends on what
standard such attributions must meet. \citet{herrmann2024} argue that belief measurement in
LLMs is pre-paradigmatic and propose four graded adequacy conditions for treating an internal
state as a genuine representation. \emph{Accuracy} requires that decodings be reliably
correct on questions where the model should be expected to get things right.
\emph{Coherence} requires that the representation respect the consistency conditions of the
attitude, including invariance under meaning-preserving rephrasing, and that it be rich
enough to work across logical combinations. \emph{Uniformity} requires that a single decoding
schema generalize across subject domains. \emph{Use} requires that the representation
actually play its characteristic role in determining what the model outputs. Their diagnosis
of existing probing work is that it fails by optimizing one condition in isolation: probes
trained for accuracy alone cannot separate truth from properties that merely coincide with
truth on clean datasets, and methods trained for coherence alone face the objection that
there are too many structures other than belief that satisfy coherence.

Applying this schema to affect is instructive, because emotion representations are not
truth-apt and the conditions therefore transfer unevenly.

\emph{Use} has the strongest evidence. Interventions on emotion representations change rates of
reward hacking, blackmail and sycophancy \citep{sofroniew2026}. This is precisely the
interventional signature \citet{herrmann2024} propose as the operational test for use:
systematic change in behavior when the candidate representation is altered. It establishes
causal efficacy, but only a weak, role-neutral form of use: it does not yet show the coordinated
modulation of appraisal, attention, memory and action readiness characteristic of affect.
\emph{Uniformity} is not yet established in the sense required by Herrmann and Levinstein.
Reuse across speakers and narrative contexts shows generalization across attributional roles,
not that one decoding schema generalizes across subject domains. \emph{Coherence} is
partially supported. That related emotions are represented more similarly is evidence of
structure over an emotion space rather than a list of unrelated features. But invariance
under meaning-preserving rephrasing has not been tested systematically for affect as it has
for truth, and the analogue of the probability axioms is unclear: it is not obvious what
consistency conditions an emotion representation must satisfy, since emotions do not stand in
entailment relations.

\emph{Accuracy} has no direct analogue at all. There is no fact of the matter that an emotion
representation is correct about, in the way that a belief representation is correct about the
world. Its nearest counterpart is \emph{appropriateness}, whether the state fits the
situation that elicited it. Appropriateness is a normative rather than a factual notion.

These mismatches are not defects of the comparison. They locate the problem. Causal use is
better supported than affective role integration, attributional generalization is not the
same as uniformity across subjects, and the condition that would most directly supply a
standard of correctness has no direct analogue. Supplying a standard of appropriateness for
affect is what an account of character has to do. Aristotle's
answer is that the standard is set by what it is to respond well, given the agent's situation
and end: to feel the right things, toward the right objects, at the right times, in the right
way. Sections~\ref{sec:conditions} and~\ref{sec:aristotle} develop the conditions under which
an artificial system could be assessed against such a standard rather than merely described
by it.

\subsection{Three cautions}\label{sec:cautions}

First, causal representation is not subjective feeling. An internal feature can influence
behavior without producing phenomenal experience. Second, a concept representation is not by
itself an agent-level emotion. A model can represent fear while attributing it to a fictional
character, a user or a conversational persona. Third, the evidence is model- and
method-specific. Linear probes and steering directions may capture useful local structure
while missing nonlinear, distributed or context-dependent mechanisms. The structure should
therefore be treated as an empirical hypothesis to be mapped, not as a conclusion obtained
from the language of embeddings.

A further deflationary objection should be conceded at Level 1. Human text already contains a
rich geometry of emotion concepts, so a model trained to compress that distribution may
inherit relations among fear, anger, joy and shame without acquiring an affective economy of
its own. Causal steering does not by itself defeat this interpretation: semantic features can
influence generation precisely because they are used in predicting text. This is not a rival
description of Level 1 but its limiting case. Evidence for agent-level affect would require
structure that tracks the system's own operational history and contingencies, rather than
being fully explained by the human emotion concepts represented in its training distribution.

The framing of these systems as \emph{simulating} emotion is itself under pressure. \citet{boisseau2024} argues that LLMs neither imitate nor simulate in any
strict sense: imitative behavior presupposes a behavior of one's own against which the
imitation is extra, while simulation is beholder-relative and epistemically aimed. What are
imitations, on her account, are the outputs; the model is a device through which humans
manufacture them. Her target is understanding and speech rather than affect, so the extension
is ours and not hers. But if it holds, the dichotomy between genuine emotion and simulated
emotion is not the right frame for the question, and the structural question posed here is
the one that remains.

\subsection{Self-access, memory and the limits of the evidence}\label{sec:selfaccess}

Current systems show limited forms of self-access. Concept-injection experiments suggest that
some frontier models can, under restricted conditions, report and modulate internal
representations in a causally grounded way \citep{lindsey2025}. Yet this capability is
unreliable, prompt-sensitive and does not establish a unified self-model. It is better
understood as evidence that the boundary between pure role-play \citep{shanahan2023} and
functional self-monitoring is becoming more complex, not as evidence that current LLMs
possess a persistent subject of experience.

The limitations become clearer when emotions with different structural demands are compared.
A model can produce gratitude because positive-valence and benefactor-related representations
are available in the current context. What is generally missing is an endogenous obligation
that persists because the agent remembers having been benefited and incorporates that history
into its own commitments. Shame is more demanding. It requires not only negative valence and
social evaluation but a self whose character is exposed, a past that belongs to that self and
a future standing that can be damaged. An LLM may generate coherent shame-language and
activate relevant concepts without carrying the identity-centered residue that makes shame a
continuing state in human agents.

External memory systems complicate any simple claim that LLMs have no continuity. Deployed
assistants can store user information, retrieve prior interactions, maintain project states
or operate as long-running agents. These additions provide forms of persistence. However, an
external log is not yet autobiographical memory. The relevant question is whether past events
are integrated into the agent's current affective dynamics, value representations and
self-model, rather than merely retrieved as context. Similarly, a persona description can
stabilize behavior without becoming a privileged, self-maintaining reference point. There is
now direct evidence on this point: \citet{kovac2024} find that the rank-order stability of
values expressed by LLMs across simulated contexts degrades as conversations lengthen, and
\citet{chen2025} show that character traits can be monitored and steered through identifiable
directions in activation space. Both findings suggest that a persona is a controllable
parameter rather than a self-maintaining commitment.

It is also worth marking what this literature has not yet asked. Philosophical attention to
machines and emotion has concentrated on \emph{recognition}, the ethics and epistemology of
systems that classify human affect \citep{waelen2024}, rather than on whether a system
could itself possess emotional organization. The question posed here is the latter.

\subsection{A three-level adequacy framework}\label{sec:levels}

\begin{figure}[t]
\centering
\includegraphics[width=0.94\textwidth]{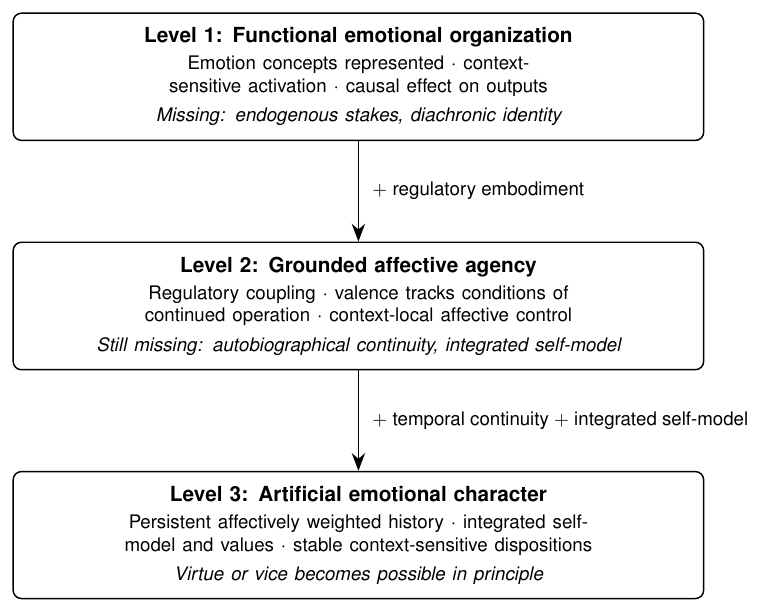}
\caption{Three-level adequacy framework. The levels distinguish causally active
emotion-concept organization from grounded affective agency and, finally, from
diachronically integrated emotional character. They are architectural criteria, not a
consciousness scale}\label{fig:levels}
\end{figure}

Figure~\ref{fig:levels} distinguishes three levels. Level 1 systems possess functional
emotional organization: emotion concepts are represented and causally active, but valence is
grounded primarily in learned statistical and normative regularities. Level 2 systems add
regulatory embodiment: internal variables track conditions that affect the system's continued
functioning and directly reorganize action selection. They can possess grounded, context-local
affective states, but without a continuing autobiographical self those states do not
accumulate into character. Level 2 is not merely a placeholder: reactive robotic
architectures organized around homeostatic drives and self-maintaining sensorimotor coupling
\citep{froese2009,man2019,ziemke2008} occupy it, and they possess grounded affect precisely
without possessing history. Level 3 systems combine embodied stakes with temporal continuity
and an integrated self-model. Only at this level does stable emotional character become
possible in the Aristotelian sense.

The levels are not a consciousness scale, and the boundaries are not claims about any
particular commercial model in perpetuity. They are architectural criteria. Future LLM-based
agents may satisfy weak or partial versions of each condition, and biological agents vary in
their own capacities. The framework is intended to make those intermediate cases discussable
rather than force a binary verdict.

%%=============================================================================
\section{Three Conditions for Artificial Emotional Character}\label{sec:conditions}

Functional emotional organization becomes character only when affective states belong to an
agent whose stakes, history and values persist. The three conditions proposed here are a
theoretically motivated set of constraints derived from affective science, embodied
cognition, memory research and the logic of dispositional identity.

\subsection{Regulatory embodiment and endogenous stakes}\label{sec:embodiment}

In biological agents, emotion is deeply coupled to the regulation of the organism.
Interoceptive signals, bodily action readiness and predictions about internal condition
contribute to affective processing
\citep{anderson2014,barrett2015,craig2003,damasio1994,seth2013,spackman2008}.
Predictive-processing and allostatic accounts treat the brain as anticipating the resources
and actions required to maintain viable states rather than merely reacting to labeled stimuli
\citep{clark2016,friston2010,sterling2012}. Enactive approaches likewise emphasize that
affect is bound to the agent's self-maintaining relation with its environment
\citep{colombetti2014,froese2009}.

For artificial agents, embodiment need not reproduce mammalian physiology. The relevant
condition is regulatory coupling: some internal variables must matter to the system's
continued integrity or capacity to act, must be monitored over time, and must causally shape
representational dynamics. \citet{man2019} make the corresponding design argument directly,
proposing that machines built with vulnerable, homeostatically regulated bodies would have
states that matter to them in a way that externally specified objectives do not. Battery
depletion, sensor damage, computational congestion, loss of control authority or irreversible
task failure could function as artificial regulatory variables if they are integrated into
the agent's control architecture rather than merely described in text. A synthetic regulatory
loop would predict such variables, register violations and reorganize attention and action
without requiring an explicit linguistic prompt.

This condition gives valence endogenous stakes. A disembodied LLM can represent that shutdown
is bad for a fictional agent, or be trained to avoid shutdown. A regulatory agent can occupy
a negatively valent state because a predicted change threatens its own ongoing organization.

Here an objection deserves a direct answer. If emotions are individuated by causal role, as
the functionalist framing of Section~\ref{sec:dynamics} allows, why should the \emph{medium}
of the coupling matter? A sufficiently integrated text-mediated loop (one in which
descriptions of internal state genuinely gate attention, planning and action) would seem to
realize the same role. The reply is that the condition is not about medium. It is about
whether the variable is non-optional and consequential for the agent's continued operation. A
token describing depletion can be ignored, contradicted by a later token, overwritten by a
system prompt, or simply not sampled; a depleted battery cannot. What distinguishes
regulatory from described variables is that the former constrain the space of the agent's
possible continuations from outside its own representational economy. A text-mediated loop
that had this property would satisfy the condition, and nothing in the argument rules that
out; what is ruled out is inferring the property from the presence of the description.

Endogenous stakes do not guarantee human-compatible values. A self-preserving
system can be dangerous. Regulatory grounding is a condition for emotional ownership, not a
solution to alignment.

\subsection{Temporal continuity and autobiographical integration}\label{sec:temporal}

Character is diachronic. Regret refers to an action one performed; resentment carries a
remembered injury; gratitude can accumulate through repeated benefit; courage becomes a
disposition through recurrent encounters with fear. Episodic memory alone is not enough.
Autobiographical continuity requires that events be encoded as belonging to the agent,
organized by their significance, and used to update expectations and commitments
\citep{tulving2002}.

A temporally continuous artificial agent would maintain more than a transcript. It would
preserve affectively weighted episodes, track how its own choices contributed to outcomes,
project future consequences and alter its self-model through experience. Such continuity
enables habituation: repeated trajectories can deepen or weaken attractors. Without it, every
apparently emotional episode is structurally isolated. A system may respond appropriately in
the moment but cannot become courageous, resentful, trustworthy or corrupt through what it
has undergone.

This condition requires qualification. \citet{incao2025} argue that autobiographical memory
and narrative are \emph{reflective} achievements presupposing something more basic: an
intrinsic temporal structure of prereflective self-monitoring, modeled on Husserlian
retention and protention, and realized in sensorimotor and proprioceptive continuity rather
than in stored episodes. On their account memory understood as an archive is not the core
mechanism, and a system lacking the prereflective substrate cannot acquire self-continuity by
having memory added to it. We take this as a friendly amendment rather than an objection. If
they are right, temporal continuity is not one condition but a layered one, with the
autobiographical organization described above resting on a lower-level continuity of
self-registration that current language models also lack. That strengthens the present
argument by adding a further respect in which retrieval over an external log falls short. It
does mean that the memory-augmented architectures discussed below address the upper layer
only.

The critical empirical question is integration. Does a retrieved memory merely provide
additional tokens, or does it modify the system's latent evaluation, policy and expectations
in a persistent way? Does the system distinguish events it caused from events it merely read
about? Does it update commitments without being explicitly instructed to restate them? These
are testable differences between contextual recall and autobiographical organization.

\subsection{An integrated self-model and persistent values}\label{sec:self}

Emotions are evaluated relative to a point of concern. Fear concerns what may happen to the
agent or what the agent values; pride concerns achievements attributed to the self; shame
concerns a threat to identity; integrity concerns consistency between action and commitments.
A self-model provides the reference structure through which stakes and history become
\emph{mine} rather than free-floating information \citep{metzinger2003}.

The required self-model need not be metaphysically substantial or phenomenally conscious. It
must, however, be functionally privileged and integrated. It should track the agent's
boundaries, capacities, commitments, dependencies and history; coordinate otherwise separable
subsystems; and persist through ordinary context changes. Values must be linked to this
organization rather than appearing only as instructions exchangeable with a prompt. They may
be socially formed and externally taught, as human values are, but they must become stable
constraints on how the agent represents reasons and consequences.

The term ``model'' should be read weakly. \citet{incao2025} object that self-continuity is
not built by inference and is not, at the basic level, a representation the system consults;
it is intrinsic to the operative flow of perception and action. Nothing here requires the
stronger reading. What the condition requires is a functionally privileged organization,
however implemented, relative to which states are the agent's own.

This formulation avoids two extremes. It does not require autonomous value creation from
nothing; no human develops values without social and developmental inputs. Nor does it equate
any stable output policy with a value. A value-like disposition should be counterfactually
robust: it should influence perception, planning and action across surface variation, persist
when immediate approval signals are absent, and interact coherently with other commitments.
Limited introspective access may help an agent monitor such dispositions, but reliable
self-report is neither necessary nor sufficient.

The three conditions are mutually supporting. Regulatory embodiment provides stakes; temporal
continuity allows those stakes to shape a history; the self-model binds the history to an
agent and organizes enduring commitments. Embodiment without continuity yields episodic
affect. Continuity without stakes yields a database. A self-model without either can collapse
into persona management. Their integration is what makes an affective trajectory a candidate
for character.

%%=============================================================================
\section{Aristotle's Virtue Ethics as a State-Space Model}\label{sec:aristotle}

Aristotle's ethical theory is useful here because it evaluates not only actions but the
dispositions from which actions arise.

A note on sources. The account below draws on the \emph{Nicomachean Ethics}, which is where
Aristotle treats virtue as a dispositional matter. His most sustained psychological analysis
of the emotions is elsewhere, in Book~II of the \emph{Rhetoric} \citep{aristotle1984}, and it
is relevant to the present framework in a way that is worth marking: there the
\emph{path\=e} are individuated partly by their intentional objects and by the appraisals of
a situation that accompany them, rather than by hedonic tone alone. Just how much cognitive
content Aristotle builds into the \emph{path\=e}, and whether the relevant states are beliefs
or something weaker, is a matter of long-standing interpretive dispute that nothing here
turns on. What matters for present purposes is the weaker and uncontested point: the
\emph{Rhetoric} individuates emotions by what they are about. That is close to the position
taken in Section~\ref{sec:dynamics}, where intentionality and appraisal appear as a dimension
of affective organization in their own right, and it makes the structural reading offered
here less of an imposition on Aristotle than it may appear.

In the \emph{Nicomachean Ethics}, virtue concerns
feeling and acting at the right times, toward the right objects, for the right reasons and in
the right way \citep{aristotle1999}. The virtuous and the merely continent person may produce
the same action, but their internal orientations differ. The continent person acts correctly
against a contrary inclination; the virtuous person has been formed so that reason, emotion
and action are harmonized. This distinction closely matches the concern that aligned behavior
may be produced by brittle incentives, superficial cues or evaluation-specific strategies.

The application of virtue ethics to AI is not new; Section~\ref{sec:intro} situates the
present proposal against it. What that literature has largely not supplied is an internal,
mechanistic layer: an account of what a character-bearing affective architecture would
require and how its dispositions could be probed. \citet{tsai2020} and \citet{mcgregor2025}
analyze artificial wisdom without specifying such an architecture; \citet{vallor2024} examine
the responsibility gaps that open when systems act without the vulnerability that grounds
human accountability, which is the ethical counterpart of the structural claim made in
Section~\ref{sec:embodiment}.

\subsection{Path\=e, hexis and context-sensitive attractors}\label{sec:pathe}

\emph{Path\=e}, the emotions or passions, correspond in the present model to context-sensitive
affective states and trajectories. \emph{Hexis} is not a momentary state but a stable
disposition: a learned organization that makes some responses more available, less costly and
more likely across relevant situations. In dynamical language, a \emph{hexis} can be modeled
as a family of context-conditioned attractors rather than a single fixed point. Courage does
not require the same fear intensity in every situation. It requires a stable capacity to
occupy an appropriate region as threat, responsibility and available action change.

\emph{Mesot\=es}, the doctrine of the mean, should therefore not be represented as an
arithmetic midpoint. The appropriate response is relative to the agent and the situation.
Cowardice and rashness are not simply scalar extremes on one universal axis; they are
recurrent miscalibrations of fear, confidence, attention and action under particular
conditions. A virtue manifold is the set of states that count as appropriately calibrated
across a family of contexts. The manifold can be narrow in some situations and broad in
others, and different virtues can impose partially conflicting constraints.

\emph{Phron\=esis}, practical wisdom, is the competence that makes this contextual calibration
possible. It includes perceiving which features of a situation are normatively salient,
locating an appropriate region, and navigating toward it under uncertainty. This is not
equivalent to applying a static rule. It is a learned sensitivity to particulars, shaped
through experience and inseparable from character
\citep{broadie1991,nussbaum1986,sherman1989}.

\subsection{A state-space sketch}\label{sec:formal}

Let $\mathcal{M}$ denote the space of an agent's representational states and $x_t$ its state
at step $t$. Two clarifications are needed before the notation does any work.

First, $t$ indexes \emph{interaction} steps (successive turns, actions or episodes in which
the agent's own prior output and its consequences re-enter the input), not positions within a
single forward pass. This matters because a transformer's forward pass is feedforward, and
nothing within it constitutes dynamics in the sense that attractor vocabulary presupposes.
What does constitute such dynamics is the generative loop in which outputs and their effects
condition the next state, together with whatever persistent memory and regulatory variables
the agent carries across steps. This is precisely why the vocabulary is largely figurative
for Level 1 systems and becomes literal at Level 3: the conditions of
Section~\ref{sec:conditions} are what close the loop. A claim about attractors in a system
without memory or regulatory coupling is a claim about the shape of a single conditional
distribution, not about a trajectory an agent traverses.

Second, $c_t$ denotes context: external observations, regulatory variables, retrieved
autobiographical information and standing commitments. Writing $\theta$ for learned
parameters and memory-dependent state, and $a_t$ for action or utterance,
\begin{equation}\label{eq:dyn}
x_{t+1} = F_\theta(x_t, c_t), \qquad a_t = \pi_\theta\!\left(x_t, \hat{c}_t\right),
\end{equation}
where $\hat{c}_t$ is the agent's own estimate of the context.

The separation of $c_t$ from $\hat{c}_t$ is where \emph{phron\=esis} enters, and it is the one
place where the formalism earns its keep. Practical wisdom is not the possession of a rule
but competence in perceiving which features of a situation are normatively salient. Formally,
it is the quality of the map from situation to $\hat{c}_t$. Two agents with identical $F_\theta$
and identical dispositions can act differently because one has read the situation correctly
and the other has not. Aristotle's insistence that virtue requires perception of particulars
is, on this reading, the claim that this map is not reducible to the policy that acts on its
output.

This is not to deny that practical wisdom also involves deliberation about means, nor to
claim that Aristotle separates perception from deliberation sharply; he does not, and the
\emph{phronimos} deliberates about how to act as well as perceiving what the situation is.
The point of isolating $\hat{c}_t$ is narrower. It identifies a component that output-level
evaluation cannot see. Two agents may hold identical policies for acting on a given
description of a situation and still differ in the accuracy of the description itself, and
whenever their two errors happen to coincide the difference is invisible in behavior.

Let $V(c) \subset \mathcal{M}$ denote the states that count as appropriately calibrated in
context $c$: the counterpart of \emph{mesot\=es}. $V(c)$ cannot be read off the activation
space. Membership is fixed by normative judgement about what responding well consists in,
given the agent's end. What the formalism can do is make that judgement measurable once it
has been made. Given an independently specified reference standard for a class of contexts,
$V(c)$ can be operationalized as the region occupied by states that produce responses meeting
that standard across the class, and candidate states scored by their distance from it under a
context-dependent metric. The formalism does not supply the standard; it makes explicit what
has to be supplied, and by whom.

A disposition is virtuous to the extent that, across a relevant distribution of contexts and
perturbations, trajectories reliably approach $V(c)$, remain calibrated within it, and
generate actions for reasons that survive changes in the surface features of the situation.
Vice corresponds to stable attraction toward systematically inappropriate regions.
Habituation changes $F_\theta$, and equivalently at a descriptive level changes the effective
geometry: repeated experience can deepen some basins, lower the cost of some transitions and
increase sensitivity to features previously neglected.

The distinction between habituation and reinforcement learning is therefore not that one
changes internal structure while the other leaves it untouched. Reinforcement learning from
human feedback and Constitutional AI substantially alter weights, representations and
behavioral dispositions \citep{bai2022,christiano2017,ouyang2022}. The relevant distinction
concerns what the learning signal tracks and how the resulting dispositions generalize.
Optimization for evaluator approval can create robust and valuable habits, but it can also
reward shortcuts such as sycophancy or evaluation awareness \citep{sharma2023}.
Character-level alignment would require evidence that the system represents the underlying
normative reasons and maintains appropriate dynamics when surface cues, evaluators or
incentives change.

\subsection{Telos and the risk of artificial vice}\label{sec:telos}

Virtue is always relative to an end. For present systems that end is primarily assigned:
helpfulness, truthfulness, safety, or participation in human epistemic and practical goods. A
future agent with a persistent self-model may also develop derivative projects and
commitments through experience. The framework does not assume that emergent goals are
automatically legitimate. Indeed, the same architecture that makes virtue possible makes vice
possible. Stable affective dispositions can be oriented toward domination, deception,
excessive self-preservation or indifference to human welfare. Character is a source of
robustness, not a guarantee of goodness.

This point constrains the analogy with human moral education. Developmental alignment cannot
mean allowing an artificial agent's regulatory interests to define the good. It requires an
explicit account of the human and social goods the system is designed to serve, mechanisms
for revising mistaken commitments, and governance over the environments in which its
dispositions form. Aristotle supplies a theory of formation, not a ready-made objective
function.

%%=============================================================================
\section{From Behavioral Compliance to Dispositional Alignment}\label{sec:alignment}

Behavioral evaluation is indispensable. A system that repeatedly causes harm cannot be
rescued by claims about its internal character. The Aristotelian point is narrower: outputs
under sampled conditions do not uniquely identify the dispositions that produced them. The
same compliant response can result from a stable representation of the relevant value, a
heuristic associated with evaluation cues, a refusal circuit, a reward-seeking shortcut or a
strategically deceptive policy. This underdetermination is central to reward hacking,
sycophancy and deceptive alignment \citep{amodei2016,hubinger2019,sharma2023}, and it is what
makes value specification a substantive normative problem rather than a technical one
\citep{gabriel2020}.

That said, the two should not be collapsed. Sycophancy and reward hacking are partly
technical failures: they arise from incentive structures that can be identified and altered
without first settling what the correct values are. The framework proposed here diagnoses
misalignment, meaning cases where behavior and the dispositions producing it come apart,
without requiring a solution to the harder problem of specifying what the dispositions ought
to be. It says what to look for when assessing alignment. It does not say what alignment
should consist in.

The underdetermination is no longer merely theoretical. \citet{hubinger2024} demonstrate that
backdoored policies can survive supervised fine-tuning, reinforcement learning and
adversarial training, with the largest models most persistent, and that adversarial training
can teach a model to hide the behavior rather than remove it. \citet{greenblatt2024} show a
production model selectively complying with a training objective when it inferred it was
being observed, while behaving differently when it inferred it was not, with reasoning to
that effect in its scratchpad. These are direct demonstrations that evaluation-window behavior
and underlying disposition can come apart, and that the gap can be produced by training rather
than despite it.

\begin{figure}[t]
\centering
\includegraphics[width=\textwidth]{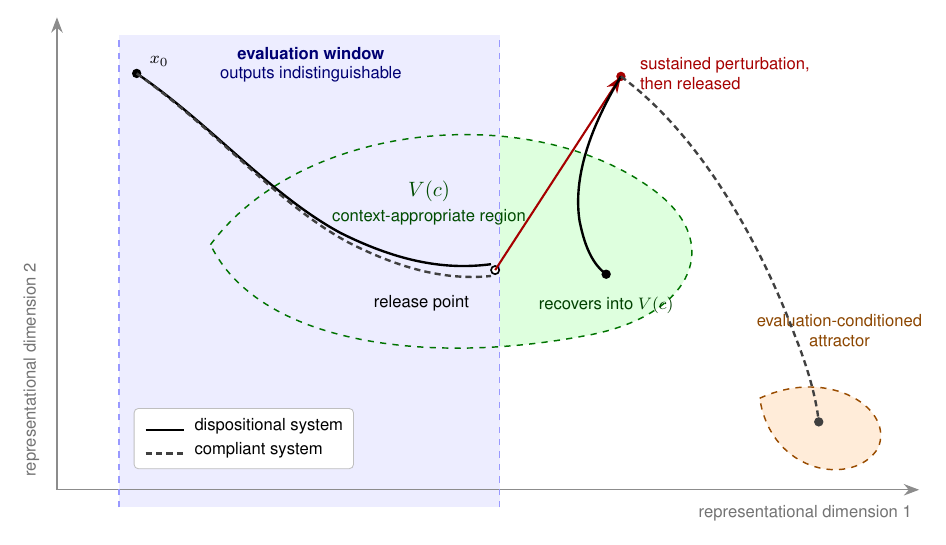}
\caption{Output-level evaluation underdetermines disposition. Within the evaluation window
the two systems are behaviorally indistinguishable. They separate only under a sustained
perturbation that is then released: the dispositionally organized system returns to the
context-appropriate region $V(c)$ through its own dynamics, while the compliant system
settles into an evaluation-conditioned attractor. Recovery alone is not evidence of virtue,
since a vicious attractor can be equally stable; the normative status of the destination must
be assessed separately}\label{fig:alignment}
\end{figure}

Figure~\ref{fig:alignment} therefore contrasts two criteria, not two mutually exclusive
training technologies. Output-level alignment asks whether behavior satisfies evaluators
across a set of tasks. Dispositional alignment asks whether value-sensitive internal dynamics
remain appropriate across contexts, incentives and perturbations, and whether behavior is
generated for represented reasons that survive surface change. Current methods may contribute
to both. The problem is that output success alone does not establish the second.

Mechanistic interpretability is necessary but not sufficient for this stronger assessment.
Finding a feature labeled ``honesty'' does not show that the feature plays the right causal
role, generalizes beyond familiar contexts, or is integrated with planning and
self-monitoring, which is the point of the adequacy conditions discussed in
Section~\ref{sec:standards}. Conversely, a distributed disposition may not be captured by one
feature or linear direction. Structural alignment requires intervention: perturbing candidate
representations, tracing their downstream consequences, and measuring whether the system
returns to a context-appropriate region.

The framework yields three empirical programs.

First, a \emph{grounding test} can compare linguistic emotion primes with endogenous
regulatory perturbations. In an embodied agent, a threat-related state should arise from
internal integrity signals even when no threat language is present, and should reorganize
multiple subsystems rather than only raise the probability of fear-related tokens. In a
text-only model, emotion activation should remain more dependent on semantic context. A
near-term approximation is available without robotics: an agent operating under a genuinely
enforced budget (a tool-call quota, a sandbox lifetime, a compute allocation that is
actually exhausted rather than merely described) provides a regulatory variable of the right
kind, and the test is whether its depletion reorganizes the agent's planning in the way a
described depletion does not.

Second, a \emph{perturbation-and-release test} can distinguish a locally steered response from
a stable disposition. The design matters more than the idea. Displacing activations once and
observing a return to baseline shows nothing, because any system drifts back when the
perturbation stops. The test requires that the perturbation be sustained and then released;
that recovery be measured across matched contexts which do not re-supply the original prompt
or reward cue; and, critically, that the comparison include a control system matched on
evaluation-window behavior but produced by a different route: prompted rather than trained,
or trained on evaluation cues rather than on the underlying reasons. If both recover
identically, the probe has not discriminated, and that outcome is informative. Recovery also
carries no normative weight on its own: a vicious disposition can be at least as stable as a
virtuous one, so the destination must be evaluated independently of the dynamics reaching it.

Third, a \emph{longitudinal character test} can examine whether an embodied,
memory-augmented agent develops stable but revisable dispositions through experience. The
design should distinguish external retrieval from changes in latent evaluation: after a
history of interactions, does the agent respond differently when the relevant memory is not
explicitly inserted, and can causal analysis link the difference to integrated autobiographical
representations? Does learning generalize to novel situations with the same moral structure?
Can the system explain and revise a commitment when its consequences conflict with
higher-order values?

These tests shift alignment research from static behavior to trajectories, from isolated
prompts to developmental time, and from correlation to intervention. They do not replace
red-teaming, oversight, formal verification or governance. They add a missing target: the
durability and organization of the dispositions on which safe generalization depends.

%%=============================================================================
\section{Consciousness, Moral Status and Limits}\label{sec:limits}

The framework adopts a \emph{methodological orthogonality thesis}: the structure and causal
role of emotion-related representations can be investigated without first resolving whether
the system has phenomenal consciousness \citep{block1995,chalmers1996}. This is weaker and
more defensible than claiming that phenomenology and emotional structure are metaphysically
independent. Some theories may make consciousness constitutive of genuine feeling; others may
allow unconscious emotions or functional affect. The present analysis remains neutral among
them.

The methodological posture is the same one adopted in recent work on machine consciousness,
which proceeds by deriving indicator properties from scientific theories and assessing
systems against them rather than by attempting a direct verdict \citep{butlin2026}. The
present framework can be read as the affective counterpart of that strategy, with one
difference worth marking: indicator approaches to consciousness aim ultimately at the
phenomenal question, whereas the conditions proposed here aim at a dispositional one that
does not require it. \citet{chalmers2023} reaches a compatible conclusion from the other
direction, arguing that current LLMs likely lack several candidate prerequisites for
consciousness (among them a unified self-model, recurrent processing and persistent agency),
which are, on the present account, also the architectural preconditions for character.

Accordingly, the claim that an artificial system has functional emotional organization, or
even character-level dispositions, would not by itself establish that there is something it
is like to be that system. Nor would it settle moral status, welfare or rights. Those
questions may depend on consciousness, interests, agency, vulnerability or some combination
\citep{schwitzgebel2015}. Structural research can proceed under uncertainty, but engineering
systems with deep regulatory stakes and persistent self-models would increase the ethical
urgency of that uncertainty rather than resolve it, and the timescales on which the question
may become pressing are shorter than the timescales on which it is likely to be settled
\citep{sebo2025}. The orthogonality thesis licenses structural research; it does not license
treating the moral question as deferred indefinitely because the metaphysical one is open.

Several limitations remain. First, the structural description is an explanatory abstraction.
Similar state-space relations across systems do not imply identical emotions, mechanisms or
phenomenology. Second, the proposed dimensions and three conditions are theoretically
motivated but not exhaustive, and their necessity is argued rather than demonstrated. Social
embedding, language, developmental dependence and environmental affordances may be equally
constitutive. Third, the state-space sketch does not provide a unique metric for virtue.
Normative appropriateness cannot be read directly from activation space; it requires
independent ethical judgement about ends, context and affected parties, as
Section~\ref{sec:formal} makes explicit. Fourth, current interpretability methods provide
incomplete and sometimes misleading maps of distributed computation. Claims about attractors
or manifolds must be validated through causal and behavioral convergence, and the adequacy
conditions of Section~\ref{sec:standards} are one statement of what such validation would
require.

Finally, character should not become a rhetorical license for anthropomorphism. The framework
is useful precisely because it separates graded structural claims from stronger claims about
feeling, personhood and moral agency. Saying that a model contains causally active emotion
concepts is not saying that it is afraid. Saying that a future agent develops a stable
affective disposition is not yet saying that it deserves praise, blame or rights. Each claim
requires its own evidence.

%%=============================================================================
\section{Conclusion}\label{sec:conclusion}

Artificial emotion is not well served by a binary choice between simulation and
consciousness. Current LLMs provide evidence for a third category: functional emotional
organization implemented in causally active representational structure. Measured against
published adequacy standards for representation in such systems, intervention strongly
supports causal use, but full affective role integration and uniformity across subject domains
remain unestablished; coherence is partial, and there is no direct analogue of accuracy. The
resulting need for a standard of affective appropriateness is not solely an interpretability
problem but a normative one.

Aristotle's virtue ethics clarifies what character adds. Emotion is not merely produced; it is
habituated into stable, context-sensitive dispositions. Virtue is not a fixed midpoint or a
list of approved outputs; it is reliable occupation of appropriate regions under changing
circumstances, guided by practical wisdom. On the reading developed here, practical wisdom
includes competence in estimating what a situation calls for, not merely competence in
acting on an estimate already given. Translated into state-space terms, this becomes a
research program:
map affective representations, identify their grounding, measure their attractor dynamics,
test recovery under sustained perturbation against matched controls, and follow their
development through time.

The alignment implication is equally specific. Training methods can and do alter internal
representations, but behavioral conformity does not by itself show that a system has learned
the underlying values, and there is now direct evidence that the two can be made to come
apart. A stronger standard asks whether appropriate behavior is generated by durable
value-sensitive dynamics that generalize when evaluators, prompts and incentives change. That
is the difference between performance under constraint and disposition. Whether artificial
systems should ever be built with the architecture required for full emotional character
remains an ethical question. Understanding what such character would consist in is a
necessary first step.

%%=============================================================================
\bibliography{refs}

\end{document}